\documentclass[letterpaper]{article} 
\usepackage{aaai24}  
\usepackage{times}  
\usepackage{helvet}  
\usepackage{courier}  
\usepackage[hyphens]{url}  
\usepackage{graphicx} 
\usepackage{natbib}  
\usepackage{caption} 
\usepackage{algorithm}

\usepackage{newfloat}
\usepackage{listings}
\DeclareCaptionStyle{ruled}{labelfont=normalfont,labelsep=colon,strut=off} 
\floatstyle{ruled}
\newfloat{listing}{tb}{lst}{}
\floatname{listing}{Listing}
\usepackage{amsfonts}       
\usepackage{amsmath}        
\usepackage{subcaption}     
\usepackage{makecell}
\usepackage{tabularx}
\usepackage{xspace}
\usepackage{booktabs}       
\usepackage{algpseudocode}
\usepackage{color} 
\usepackage{hyperref} 

\title{Not All Tasks Are Equally Difficult: \\ Multi-Task Deep Reinforcement Learning with Dynamic Depth Routing}
\author{
    Jinmin He\textsuperscript{\rm 1,2},
    Kai Li\textsuperscript{\rm 1,2}\thanks{Corresponding authors.},
    Yifan Zang\textsuperscript{\rm 1,2},
    Haobo Fu\textsuperscript{\rm 6},
    Qiang Fu\textsuperscript{\rm 6},
    Junliang Xing\textsuperscript{\rm 5}\footnotemark[1],
    Jian Cheng\textsuperscript{\rm 1,3,4}
}
\affiliations{
    \textsuperscript{\rm 1}Institute of Automation, Chinese Academy of Sciences\\
    \textsuperscript{\rm 2}School of Artificial Intelligence, University of Chinese Academy of Sciences\\
    \textsuperscript{\rm 3}School of Future Technology, University of Chinese Academy of Sciences\\
    \textsuperscript{\rm 4}AiRiA \\
    \textsuperscript{\rm 5}Tsinghua University \\
    \textsuperscript{\rm 6}Tencent AI Lab
    
    \{hejinmin2021, kai.li, zangyifan2019, jian.cheng\}@ia.ac.cn, 
    \{haobofu, leonfu\}@tencent.com, 
    jlxing@tsinghua.edu.cn
}

\makeatletter
\DeclareRobustCommand\onedot{\futurelet\@let@token\@onedot}
\def\@onedot{\ifx\@let@token.\else.\null\fi\xspace}
\def\eg{\emph{e.g}\onedot} 
\def\ie{\emph{i.e}\onedot}

\def\with{\emph{w/} }
\def\without{\emph{w/o} }

\makeatother

\begin{document}

\maketitle

\begin{abstract}
Multi-task reinforcement learning endeavors to accomplish a set of different tasks with a single policy.
To enhance data efficiency by sharing parameters across multiple tasks, a common practice segments the network into distinct modules and trains a routing network to recombine these modules into task-specific policies.
However, existing routing approaches employ a fixed number of modules for all tasks, neglecting that tasks with varying difficulties commonly require varying amounts of knowledge.
This work presents a Dynamic Depth Routing (D2R) framework, which learns strategic skipping of certain intermediate modules, thereby flexibly choosing different numbers of modules for each task.
Under this framework, we further introduce a ResRouting method to address the issue of disparate routing paths between behavior and target policies during off-policy training.
In addition, we design an automatic route-balancing mechanism to encourage continued routing exploration for unmastered tasks without disturbing the routing of mastered ones.
We conduct extensive experiments on various robotics manipulation tasks in the Meta-World benchmark, where D2R achieves state-of-the-art performance with significantly improved learning efficiency.
\end{abstract}

\section{Introduction}

Deep Reinforcement Learning (RL) has witnessed remarkable advancements in the past decade, demonstrating successful applications in diverse domains, including game playing~\cite{mnih2015human, ye2020mastering} and robotic control~\cite{lillicrap2015continuous,levine2016end}.
However, despite the current capabilities of deep RL methods to learn individual policies for specific tasks, training a single policy that can generalize across multiple tasks remains highly challenging~\cite{teh2017distral,yu2020meta}.

Multi-Task Reinforcement Learning (MTRL) aims to acquire an effective policy capable of accomplishing a diverse range of tasks.
Given the potential similarities among the joint learning tasks, training a multi-task policy allows for sharing and reusing knowledge across related tasks, resulting in enhanced sample efficiency compared to training each task separately~\cite{yang2020multi}.
However, even within the same domain, the variability between tasks still exists, making it challenging to determine what knowledge should be shared between tasks and how to share it appropriately.

\begin{figure}[t]
\centering
\includegraphics[width=0.95\columnwidth]{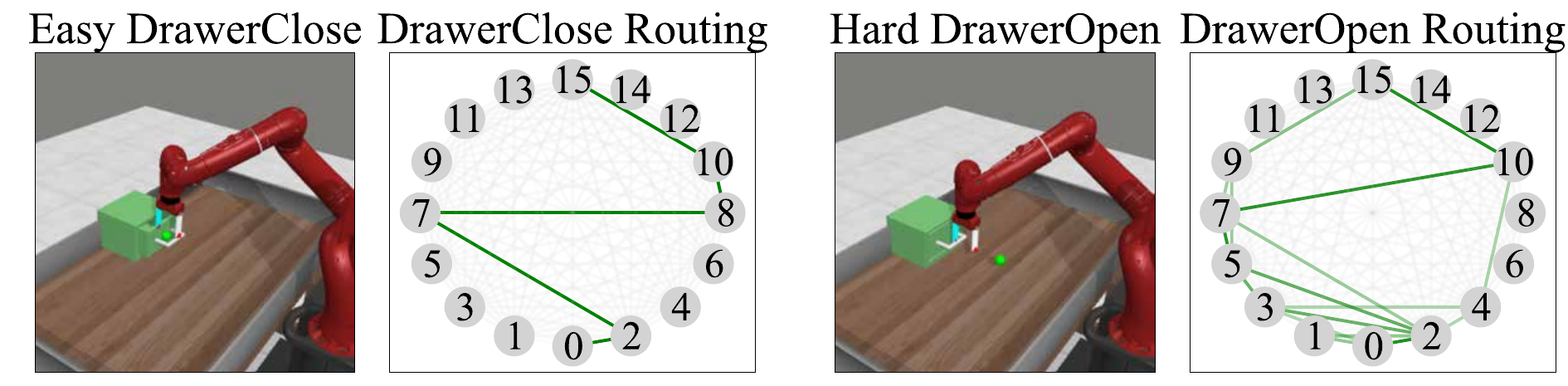}
\caption{
The easy task \textit{drawer-close} can be solved by simply pushing at any position of the drawer, so D2R learns to use only 6 modules.
However, when tackling the difficult task \textit{drawer-open}, which involves first grasping the handle and then pulling outward, D2R adapts to utilize 10 modules.
}
\label{fig:easy_difficult}
\end{figure}

To tackle this problem, a class of approaches based on modularization and routing has been proposed~\cite{fernando2017pathnet, yang2020multi, sun2022paco}.
These approaches usually consist of two networks:
a base module network, which divides all parameters into separate modules to learn different knowledge, and a routing network, which determines how these modules are combined.
These two networks are jointly optimized over multiple tasks.

\begin{figure*}[t]
\centering
\begin{subfigure}[t]{0.23\textwidth}
  \centering
  \includegraphics[width=0.95\columnwidth]{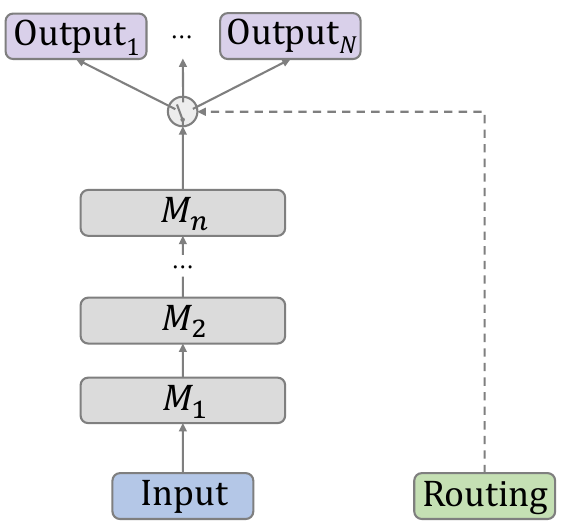}
  \caption{Output-level Routing}
\end{subfigure}
\begin{subfigure}[t]{0.23\textwidth}
  \centering
  \includegraphics[width=0.95\columnwidth]{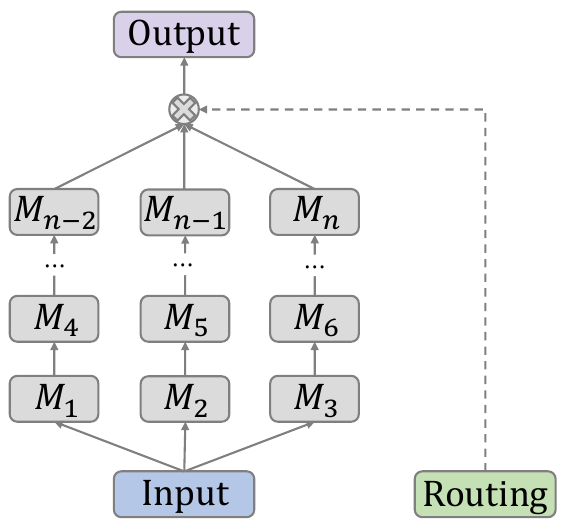}
  \caption{Model-level Routing}
\end{subfigure}
\begin{subfigure}[t]{0.23\textwidth}
  \centering
  \includegraphics[width=0.95\columnwidth]{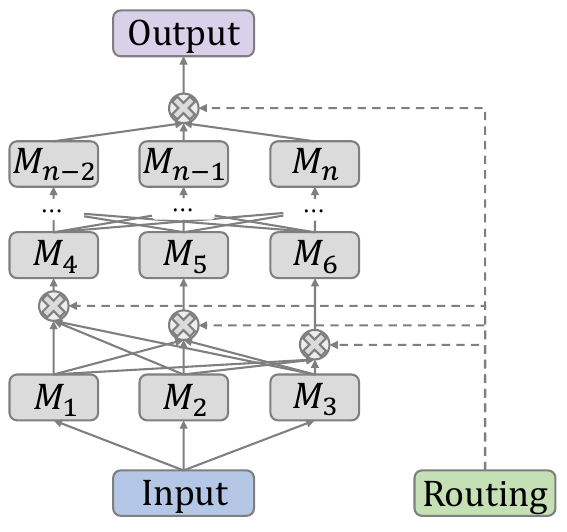}
  \caption{Layer-level Routing}
\end{subfigure}
\begin{subfigure}[t]{0.23\textwidth}
  \centering
  \includegraphics[width=0.95\columnwidth]{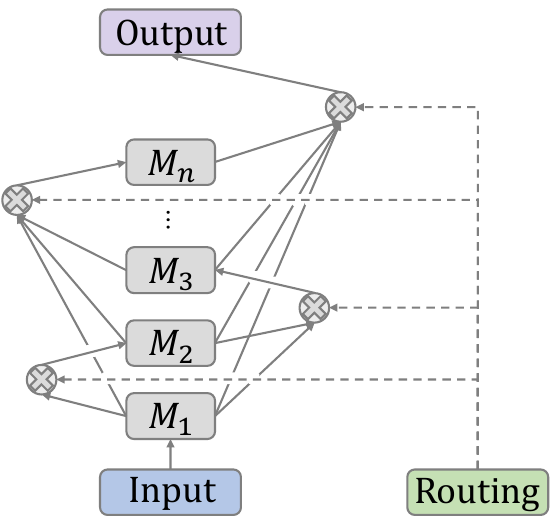}
  \caption{Module-level Routing}
\end{subfigure}
\caption{Four different levels of routing with $n$ modules.
(a) is the basic multi-head approach.
(b) routes from several separate networks.
(c) establishes connections between adjacent layers.
(d) can build any possible connections to form a DAG routing.}
\label{fig:module-level}
\end{figure*}

Although these existing approaches are promising, their routing networks always use the same number of modules for all tasks.
It could result in unnecessary parameter optimization because \textit{not all tasks are equally difficult}.
Different tasks demand varying amounts of knowledge, with easier tasks typically requiring fewer modules than difficult ones.
Based on this consideration, we introduce a Dynamic Depth Routing (D2R) framework that
combines a diverse number of modules with flexible adaptation to task difficulty (Fig.\ref{fig:easy_difficult}).
Specifically, the modules within D2R's base module network are arranged in a predetermined topological sequence.
For each task, each module establishes routing connections with selected preceding modules.
Thus a module is skipped in case any of its following modules does not select it.
In this way, D2R automatically determines which modules to use and the number of modules required for each task.

Our D2R framework is agnostic to the underlying RL algorithms.
However, during off-policy training, it faces a unique challenge of disparate routing paths between behavior and target policies.
To address this issue, we propose a ResRouting method, which avoids potential negative knowledge transfer by prioritizing updating the behavior policy routing without optimizing unsuitable modules for the target policy.
In addition, D2R also suffers from the prevalent imbalanced learning problem during multi-task training, wherein easy tasks typically converge faster than difficult ones.
To address this problem, we propose an automatic route-balancing mechanism to harmonize the learning process of different tasks by encouraging continued routing exploration for unmastered tasks while simultaneously maintaining the routing stability for already mastered tasks.

We conduct extensive experiments in the Meta-World benchmark~\cite{yu2020meta}, consisting of 50 robotic manipulation tasks.
D2R achieves significant improvement in both sample efficiency and final performance compared to state-of-the-art MTRL algorithms.
We perform rich analyses to demonstrate that D2R learns suitable routing paths based on task difficulty.
We also conduct detailed ablation studies to gain insights into the contribution of each component.

\section{Background}

\subsection{Multi-Task Reinforcement Learning}
We consider MTRL setting involving $N$ tasks, with each task $\mathcal{T}$ represented by a finite horizon Markov Decision Process (MDP) \cite{bellman1966dynamic, puterman2014markov}. Each MDP can be characterized by $\langle S, A, P, R, \gamma \rangle$, where states $s \in S$ and actions $a \in A$ are continuous in this paper.
In addition, $P(s_{t+1}|s_t, a_t)$ denotes the stochastic transition dynamics, $R(s_t, a_t)$ represents the reward function, and $\gamma$ is the discount factor.
In the context of MTRL, the primary objective is to learn a policy $\pi(a_t|s_t)$ that maximizes the expected return across all tasks sampled from a task distribution $p(\mathcal{T})$.

\subsection{Soft Actor-Critic}
In this work, we use the Soft Actor-Critic (SAC)~\cite{haarnoja2018soft} algorithm to train a policy across multiple tasks.
SAC is an off-policy actor-critic method that leverages the maximum entropy framework.
The critic network $Q_\theta(s_t, a_t)$, representing a soft Q-function, is parameterized by $\theta$, and the actor network $\pi_\phi(a_t|s_t)$ is parameterized by $\phi$.
The objective of policy optimization is:
\begin{equation}
\label{eq:sac}
\resizebox{0.9\linewidth}{!}{$
J_\pi(\phi) = \mathbb{E}_{s_t \sim \mathcal{D}} \left[ \mathbb{E}_{a_t \sim \pi_\phi} \left[ \alpha \log \pi_\phi(a_t|s_t) - Q_\theta(s_t,a_t) \right] \right],
$}
\end{equation}
where $\mathcal{D}$ represents the data in the replay buffer, and $\alpha$ is a learnable temperature parameter to penalize entropy.
It is learned to maintain the entropy level of the policy following:
\begin{equation}
\label{eq:sac_alpha}
J(\alpha) = \mathbb{E}_{a_t \sim \pi_\phi}  \left[ -\alpha \log \pi_\phi(a_t|s_t) - \alpha \bar{\mathcal{H}} \right],
\end{equation}
where $\bar{\mathcal{H}}$ is a desired minimum expected entropy.
If the optimization leads to an increase in $\pi_\phi(a_t|s_t)$ with a decrease in the entropy, the temperature $\alpha$ will accordingly increase.

\subsection{Modularization and Routing}
Training a multi-task policy with global parameter sharing poses a challenging optimization problem due to potential conflicts in gradient descent across different tasks~\cite{yu2020gradient, liu2021conflict}.
\textit{Modularization} involves dividing parameters into different modules, which reduces gradient interference by utilizing different modules and enables better generalization~\cite{singh1992transfer, andreas2017modular}.
We view module reuse as a form of knowledge sharing.
Considering that different tasks and states may necessitate varying knowledge, it is natural to require a rational selection and combination of modules, which is referred to as \textit{routing}.
To better position our approach, we categorize the existing routing frameworks in Fig.\ref{fig:module-level}(a), (b), and (c).

\textbf{Output-level routing} (Fig.\ref{fig:module-level}(a))
is the most basic and standard multi-head approach, where tasks share all the modules in the backbone, but each task has its specific head module.
It is a fixed routing with no learning process, as each head deterministically matches a specific task.

\textbf{Model-level routing} (Fig.\ref{fig:module-level}(b))
involves routing from several separate networks comprising multiple modules, which shares a similar idea with the Mixture-of-Experts (MoE) architecture \cite{jacobs1991adaptive, shazeer2017outrageously}.
In this context, the routing network learns a weight assignment for separate networks, which is analogous to the gating network in MoE used to select a combination of experts.

\textbf{Layer-level routing} (Fig.\ref{fig:module-level}(c))
is an extension of model-level routing.
It enables each layer to apply separate routing and establish routing connections between modules in two adjacent layers~\cite{yang2020multi}, but connections across layers are not allowed.
The routing network learns the combining weights of each module within each layer.

\section{Method}

The computational cost remains consistent among various tasks in all three routing types mentioned above since the amount of parameter usage remains unchanged across tasks.
However, easy tasks only require simpler policies with fewer parameters than difficult ones.
Instead of dividing modules into sets and requiring routing between specified sets, we treat each module as an independent unit and propose a new \textbf{module-level routing} (Fig.\ref{fig:module-level}(d)).
In this paradigm, a predetermined topological order organizes all modules. 
The routing network learns a separate Directed Acyclic Graph (DAG) for each task that satisfies a subsequence of this ordering.
In each DAG, the vertexes represent the utilized module, and the directed edges represent the learned routing connections between these modules.

In terms of flexibility, output-level routing is the least flexible since its routing path is static and solely determined by task identity.
In contrast, the other three routing types allow for more flexible knowledge sharing because their routing paths dynamically change during the training process, and they can incorporate additional clues, such as the current state information.
Regarding expressive capability, module-level routing stands out as the strongest. 
Both model-level and layer-level routing paths can be viewed as specific DAGs with structural restrictions.
For instance, layer-level routing prohibits cross-layer connections between non-adjacent layers.
Therefore, they are special cases of module-level routing, which does not constrain DAG's structure.

\subsection{Dynamic Depth Routing}

\begin{figure}[t]
\centering
\includegraphics[width=0.81\columnwidth]{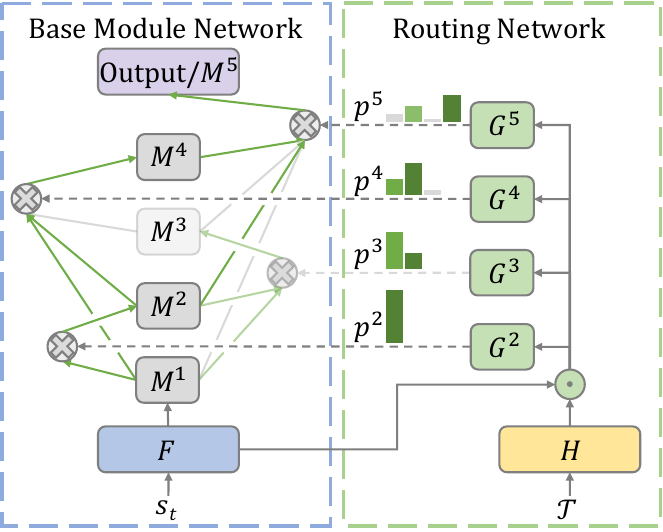}
\caption{The structure of D2R contains a base module network (left) with multiple modules and a routing network (right) that generates the routing probabilities $p^i$ for each module to select and combine its routing sources.}
\label{fig:depth_route}
\end{figure}

Considering the varying knowledge requirements for tasks of different complexities, our model aims to select appropriate modules and combine them optimally based on the task difficulty.
To achieve this goal, we introduce a Dynamic Depth Routing (D2R) framework, which builds upon module-level routing.
As shown in Fig.\ref{fig:depth_route}, D2R contains two networks: a base module network and a routing network.
In each timestep, the routing network recombines the base modules using the current observed state $s_t$ and the task $\mathcal{T}$.
In this work, we use a one-hot vector to represent each task.

\begin{figure*}[t]
\centering
\centering
\includegraphics[width=0.99\textwidth]{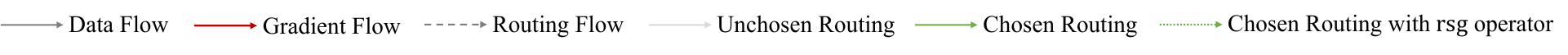}

\centering
\begin{subfigure}[b]{0.26\textwidth}
  \centering
  \includegraphics[height=92pt]{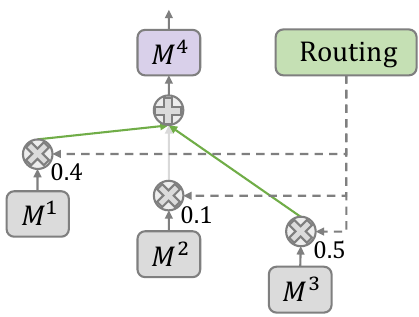}
  \caption{Behavior Policy $\pi_{\phi_{\text{old}}}$'s Routing}
\end{subfigure}
\begin{subfigure}[b]{0.27\textwidth}
  \centering
  \includegraphics[height=92pt]{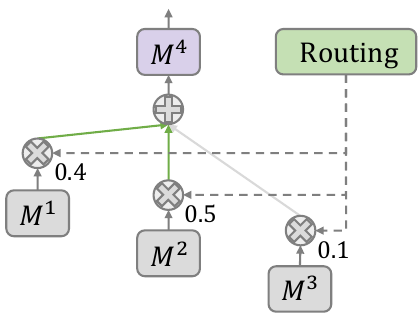}
  \caption{Target Policy $\pi_{\phi}$'s Routing}
\end{subfigure}
\begin{subfigure}[b]{0.42\textwidth}
  \centering
  \includegraphics[height=92pt]{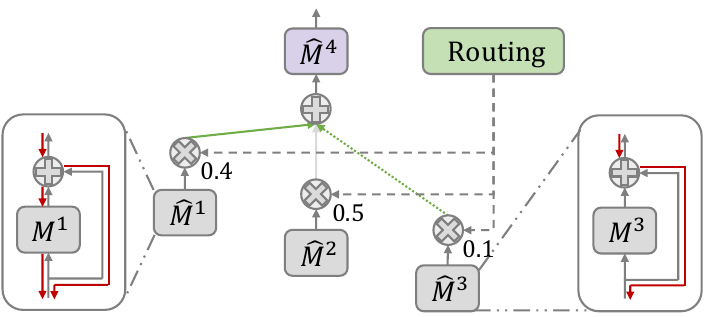}
  \caption{ResRouting for Off-policy Training}
\end{subfigure}
\caption{(a) and (b) illustrate the disparate routing paths between behavior policy and target policy during off-policy training.
(c) shows the structure of ResRouting, where routing sources with low probabilities are processed using the $\operatorname{rsg}$ operator.}
\label{fig:resrouting}
\end{figure*}

\paragraph{Base Module Network.}
Suppose we have $n$ modules in total, and all modules are arranged in a predetermined topological sequence.
The input to the first module is the state representation $F(s_t)$.
For the remaining, the input of the $i$-th module is calculated as a weighted sum of the outputs from previous modules $\{m^j \! : \!j<i\}$, determined by a routing probability $p^i$ which is generated by the routing network.
The output of the $i$-th module is represented as:
\begin{equation}
\label{eq:base_module}
m^i = M^{i}\left(\sum_{j=1}^{i-1} p^i_j \cdot m^j\right),
\end{equation}
where $M^i$ is the network of the $i$-th module.
The output dimensions are consistent across all modules except for the last one $m^n$, which should align with the action distribution.

\paragraph{Routing Network.}
The routing network consists of $n$ sub-networks, each generating routing probability $p^i$ for its corresponding base module.
The input to the routing network is the element-wise multiplication of the state representation $F(s_t)$ and the task representation $H(\mathcal{T})$.
We define the routing logit of module $i$ as $z^i \in \mathbb{R}^{i-1}$, represented as:
\begin{equation}
\label{eq:routing}
z^i = G^i \left( F(s_t)\cdot H(\mathcal{T}) \right),
\end{equation}
where $G^i$ is the routing sub-network for the $i$-th module.

Not all tasks are equally difficult, with easy tasks commonly requiring fewer modules. Therefore, not every task necessitates the full connection of all modules.
In addition, in order to train the network in an end-to-end manner, the combination process is required to be differentiable.
Thus, we use Top-K MaskSoftmax function to normalize routing logit $z^i$ into routing probability $p^i$ as:
\begin{equation}
\label{eq:maskedsoftmax}
p^i_j = \operatorname{MaskSoftmax}\left(z^i, d^i\right)_j = \frac{\exp(z^i_j) \cdot d^i_j}{\sum_{k=1}^{i-1} \exp(z^i_k) \cdot d^i_k},
\end{equation}
where $d^i$ is the routing path of module $i$ represented as:
\begin{equation}
\label{eq:topk}
d^i_j = \operatorname{TopK}\left( z^i, k \right)_j = \left\{ 
\begin{array}{ll}
    1 & \text{if} \ z^i_j \ \text{in top} \ k \ \text{of} \ z^i. \\
    0 & \text{otherwise}.
\end{array}
\right.
\end{equation}

The deterministic routing from top $k$ modules always amounts to an exploitative strategy, resulting in different tasks converging to the same routing paths.
To promote exploration and discover improved routing paths, we modify Eq.\ref{eq:topk} by sampling $k$ routing sources without replacement from the multinomial distribution in the training phase as:
\begin{equation}
\label{eq:samplek}
d^i_j = \operatorname{SampleK}\left( z^i, k \right)_j = \left\{
\begin{array}{ll}
    1 & \text{if} \ z^i_j \ \text{in sampled} \ k. \\
    0 & \text{otherwise}.
\end{array}
\right.
\end{equation}
If the number of preceding modules is less than $k$, the routing sources include all of them.
Referring to the setting of the sparse MoE model~\cite{shazeer2017outrageously, lepikhin2020gshard}, we set $k=2$ in our implementation.

As shown in Fig.\ref{fig:depth_route}, five modules $\{M^i\}_{i=1}^5$ are included, each of which gets its own routing sources indicated by the green arrows.
However, $M^3$ is actually unutilized because the routing path back from output does not go through $M^3$ (neither $M^5$ nor $M^4$ chooses routing from $M^3$).
Hence D2R strategically skips the calculation of certain base modules.

The critic network adopts a similar architecture as the actor network described above, with an additional input parameter (action $a_t$) and a different output of Q-value.

\subsection{ResRouting for Off-policy Training}
D2R is a generic framework agnostic to the underlying RL algorithms, supporting both on-policy and off-policy approaches.
As sample efficiency is a primary motivation for MTRL, off-policy algorithms are more commonly used~\cite{yu2020meta}, which learn with substantially fewer environment samples than on-policy ones.
However, D2R faces a unique challenge during off-policy training.
Specifically, the behavior policy $\pi_{\phi_{\text{old}}}$, which generates training trajectories, may follow different routing paths from the target policy $\pi_{\phi}$, which is currently evaluated and improved.
Fig.\ref{fig:resrouting} illustrates a top-2 routing sample where module $M^4$ chooses to route from $M^1$ and $M^3$ due to the highest routing probabilities in $\pi_{\phi_{\text{old}}}$ (Fig.\ref{fig:resrouting}(a)); for now, $M^1$ and $M^2$ are chosen by $\pi_{\phi}$ (Fig.\ref{fig:resrouting}(b)).
Thus, a key question is how to update the target policy under this routing path inconsistency.

We propose a ResRouting method (Fig.\ref{fig:resrouting}(c)) to address this challenge.
Given that $\pi_{\phi_{\text{old}}}$ generates actions through the routing path $d_\text{old}$ to interact with the environment, it is natural to evaluate and update the corresponding routing strategy.
Therefore, we store the routing paths $d_{\text{old}}$ of $\pi_{\phi_{\text{old}}}$ in the replay buffer and utilize them to train $\pi_{\phi}$.
The routing probability of $\pi_{\phi}$ in Eq.\ref{eq:maskedsoftmax} is modified as:
\begin{equation}
\label{eq:keepk-maskedsoftmax}
p^i_j = \operatorname{MaskSoftmax}\left(z^i, d^i_\text{old}\right)_j.
\end{equation}

However, blindly optimizing all the modules in $d_{\text{old}}$ may cause negative knowledge transfer since some modules in $d_{\text{old}}$ might no longer be suitable in $\pi_\phi$ since it has been trained for a while.
The optimization of these unsuitable modules does not yield performance improvement; on the contrary, it adversely impacts the other paths that utilize these modules.
Therefore, we prevent the update of these unsuitable modules in $\pi_\phi$.
The output of each module in Eq.\ref{eq:base_module} is modified as:
\begin{equation}
\label{eq:resrouting2}
m^i = M^{i} \left(\sum_{j=1}^{i-1} p^i_j \cdot \chi\left(m^j, z^i_j, \operatorname{sg}\right) \right),
\end{equation}
where $\operatorname{sg}$ stands for the stop-gradient operator, defined as an identity function during forward computation with zero partial derivatives.
$\chi(m^j, z^i_j, \operatorname{sg})$ is an indicator function used to decide the application of the $\operatorname{sg}$ operator on module $m^j$,
\begin{equation}
\chi\left(m^j, z^i_j, \operatorname{sg}\right) \hspace{-1pt} = \hspace{-1pt} \left\{ \hspace{-2pt}
\begin{array}{ll}
    m^j                      & \hspace{-3pt} \text{if} \ \operatorname{Softmax}(z^i)_j \ge \sigma^i, \\
    \operatorname{sg}[{m}^j] & \hspace{-3pt} \text{if} \ \operatorname{Softmax}(z^i)_j < \sigma^i,
\end{array}
\right.
\end{equation}
where threshold $\sigma^i$ relates to the number of optional routing sources, which is set to $1/i$.
As shown in Fig.\ref{fig:resrouting}, $M^3$ is unsuitable for $M^4$ when updating $\pi_{\phi}$ ($0.1 < 1/4$) even though it was previously selected by $\pi_{\phi_\text{old}}$.

However, apart from blocking the gradients of $M^3$, the original $\operatorname{sg}$ operator also prevents updating all $M^3$'s routing source modules.
Therefore, later modules are more likely to be optimized than the preceding ones.
It raises a subtle optimization problem of reduced training efficiency and unbalanced updating frequency.
Inspired by the residual learning in ResNet~\cite{he2016deep}, ResRouting adopts a similar structure with an additional shortcut connection.
For suitable modules (like $M^1$ in Fig.\ref{fig:resrouting}(c)), ResRouting preserves the full gradient information.
However, for unsuitable modules (like $M^3$ in Fig.\ref{fig:resrouting}(c)), ResRouting stops the gradient to avoid updating these specific modules while retaining the gradient information in the shortcut connection, thereby facilitating the update of preceding modules.
So Eq.\ref{eq:resrouting2} is modified as:
\begin{equation}
\label{eq:resrouting3}
m^i = \sum_{j=1}^{i-1} p^i_j \cdot \hat{m}^j + M^{i} \left(\sum_{j=1}^{i-1} p^i_j \cdot \hat{m}^j \right),
\end{equation}
where $\hat{m}^j = \chi\left(m^j, z^i_j, \operatorname{rsg}\right)$,
and $\operatorname{rsg}$ operator is defined as:
\begin{equation}
\operatorname{rsg}[m^i] = \sum_{j=1}^{i-1} p^i_j \cdot \hat{m}^j + \operatorname{sg}\left[ M^{i} \left(\sum_{j=1}^{i-1} p^i_j \cdot \hat{m}^j \right) \right].
\end{equation}

In summary, ResRouting satisfactorily prioritizes updating the routing of the behavior policy while avoiding updating unsuitable modules in the current target policy simultaneously.
Furthermore, it enables the effective training of all modules, thereby enhancing training efficiency.

\subsection{Automatic Route-Balancing}
Besides the off-policy routing inconsistency issue, D2R faces another challenge of an imbalanced learning process when simultaneously optimizing multiple tasks, where easy tasks typically converge faster than difficult ones.
To this end, we propose an automatic route-balancing mechanism to harmonize the learning process of different tasks.
It encourages continued exploration of routing paths for the unmastered difficult tasks while exploiting routing paths for the mastered easy ones.
Specifically, we introduce a hyper-parameter $\tau_\mathcal{T}$ to the routing path sampling for each task $\mathcal{T}$,
\begin{equation}
\label{eq:temp-samplek}
d^i_j = \operatorname{SampleK}\left( z^i/\tau_\mathcal{T}, k \right)_j,
\end{equation}
where $z^i/\tau_\mathcal{T}$ replaces $z^i$ in Eq.\ref{eq:samplek}.
For unmastered tasks, $\tau_\mathcal{T}$ should be larger to make the routing distribution smoother, thus facilitating exploration.
In contrast, for mastered tasks, $\tau_\mathcal{T}$ should be smaller to emphasize exploitation.

Manually specifying $\tau_\mathcal{T}$ for each task is cumbersome and time-consuming.
Inspired by \cite{yang2020multi}, we can automatically adjust $\tau_\mathcal{T}$ online by leveraging the favorable properties of the SAC temperature parameter $\alpha$ in Eq.\ref{eq:sac_alpha} as:
\begin{equation}
\label{eq:temp}
\tau_\mathcal{T} = \frac{1/\alpha_\mathcal{T}}{\sum_{j=1}^{N} \left( 1 / \alpha_j \right)},
\end{equation}
where $\alpha_\mathcal{T}$ is the SAC temperature parameters for task $\mathcal{T}$, and $N$ is the total number of tasks.
For unmastered (mastered) tasks, their policy entropy tends to be high (small). 
SAC decreases (increases) the value of $\alpha$ through Eq.\ref{eq:sac_alpha}, which, in turn, increases (decreases) the value of $\tau_\mathcal{T}$ according to Eq.\ref{eq:temp}, meeting our requirements.
The full pseudocode for D2R is described in Appendix B (Algorithm 1).

\section{Related Work}
\paragraph{Multi-task Reinforcement Learning.}
Multi-task learning is a training paradigm that improves generalization by utilizing the domain information contained in potentially related tasks \cite{caruana1997multitask}.
MTRL extends the idea of multi-task learning to reinforcement learning, expecting that knowledge shared across tasks will be discovered by simultaneously learning multiple RL tasks.
It has been extensively investigated from various perspectives \cite{calandriello2014sparse, hessel2019multi, xu2020knowledge}.
\cite{yu2020gradient, liu2021conflict} view MTRL as a multi-objective optimization problem aiming to handle conflicting gradients arising from different task losses during training, but they lead to an increase in computational and memory consumption.
\cite{parisotto2015actor, teh2017distral} utilize policy distillation to integrate separate policies for different tasks into a unified policy, yet they still require separate networks for different policies and an additional distillation phase.
\cite{sodhani2021multi, cho2022multi} attempt to choose or learn better representations as more effective task-condition for policy training.
In addition, \cite{sodhani2021multi} can be viewed as a model-level routing approach because it leverages a mixture of encoders to learn a context-dependent representation effectively.
\cite{yang2020multi} stands for layer-level routing, which generates the combination weight for modules between layers.
Instead of routing for the outputs of modules, \cite{sun2022paco} explores a compositional parameter space to reconfigure the parameters of each task's network.
All of the above routing approaches limit the number of parameters per task to remain the same.
Differently, D2R proposes a more flexible routing scheme that can reasonably allocate the number of modules among different tasks according to the task difficulty.

\paragraph{Dynamic Neural Network.} 
The classical dynamic network architecture Mixture-of-Experts (MoE) \cite{jacobs1991adaptive, ma2018modeling} builds multiple networks as experts in parallel, performing inference with dynamic width.
Sparse MoE \cite{shazeer2017outrageously, lepikhin2020gshard, fedus2022switch} inspires our work that the amount of computation can remain relatively small compared to the total model size.
Our work is more related to studies on dynamic neural networks with dynamic depth.
The early-exiting paradigm allows easy samples to be output at shallow exits without executing deeper layers \cite{teerapittayanon2016branchynet, bolukbasi2017adaptive}, but the cross-layer connections are not permitted.
Instead of skipping the execution of all the deep layers after a certain classifier, methods like \cite{wang2018skipnet, veit2018convolutional} selectively skip intermediate layers conditioned on samples.
Besides optionally skipping some modules, D2R also dynamically recombines routing sources for each module.

\section{Experiments}

\subsection{Experimental Setup}

\begin{figure*}[t]
\centering
\includegraphics[width=0.94\textwidth]{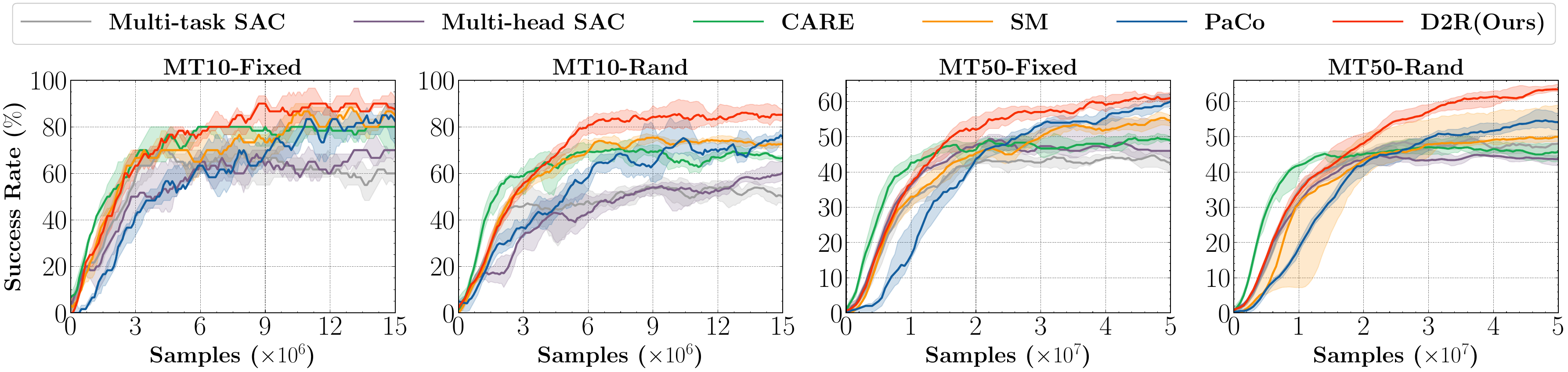}
\caption{Comparison of D2R against baselines on four benchmark settings.}
\label{fig:result}
\end{figure*}

\begin{figure*}[t]
\centering
\begin{subfigure}[b]{0.32\textwidth}
  \centering
  \includegraphics[height=108pt]{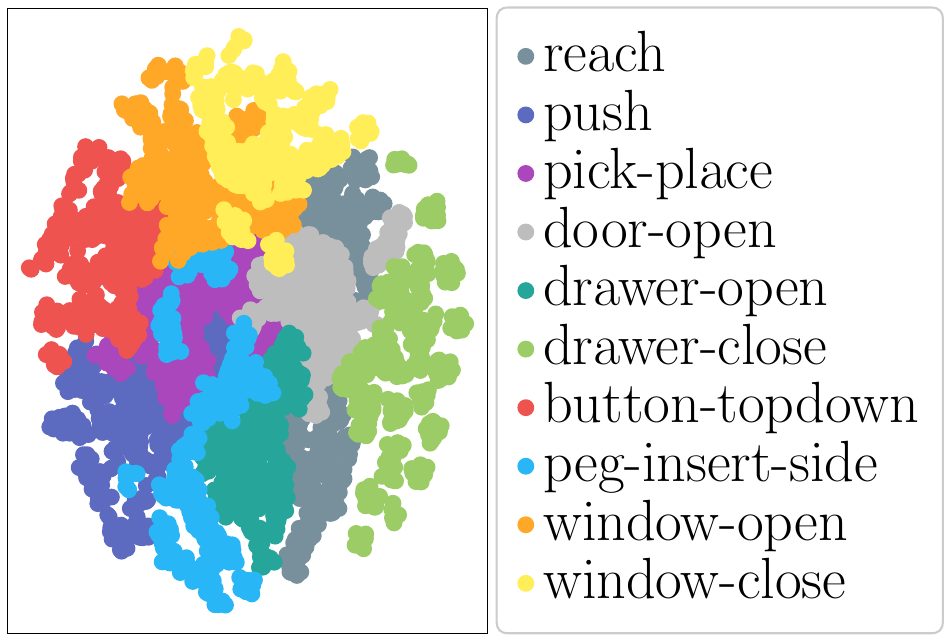}
  \caption{t-SNE Visualization}
\end{subfigure}
\begin{subfigure}[b]{0.21\textwidth}
  \centering
  \includegraphics[height=108pt]{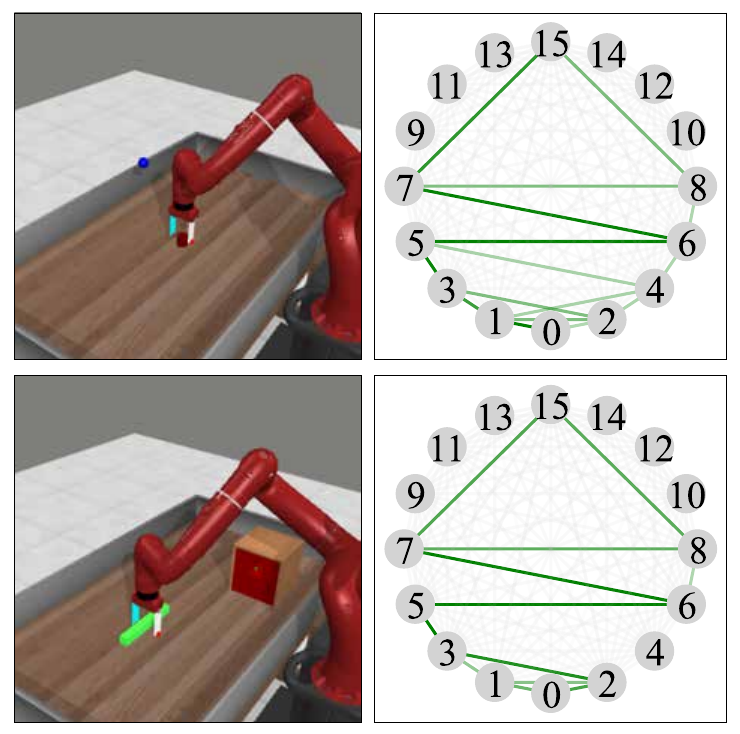}
  \caption{pick-place/peg-insert-side}
\end{subfigure}
\begin{subfigure}[b]{0.21\textwidth}
  \centering
  \includegraphics[height=108pt]{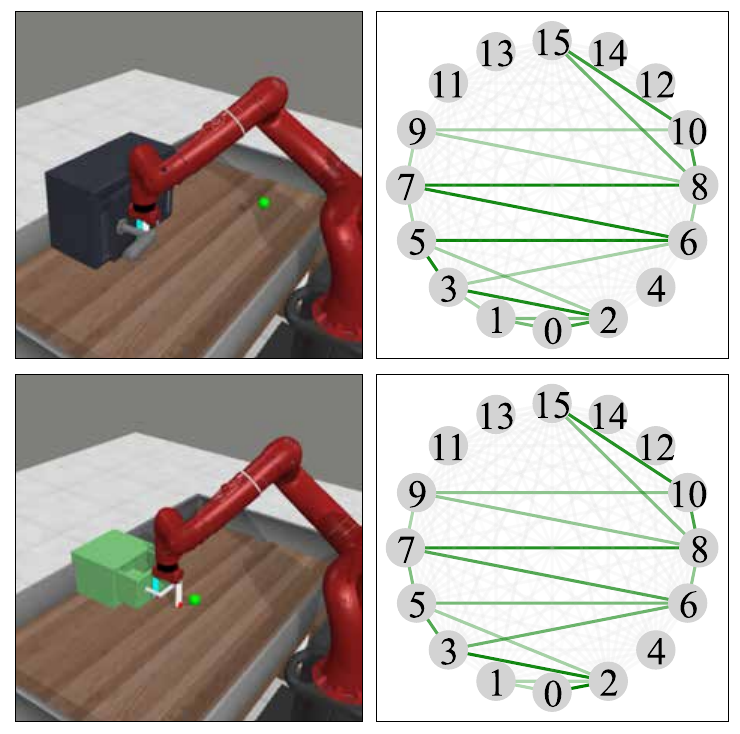}
  \caption{door/drawer-open}
\end{subfigure}
\begin{subfigure}[b]{0.21\textwidth}
  \centering
  \includegraphics[height=108pt]{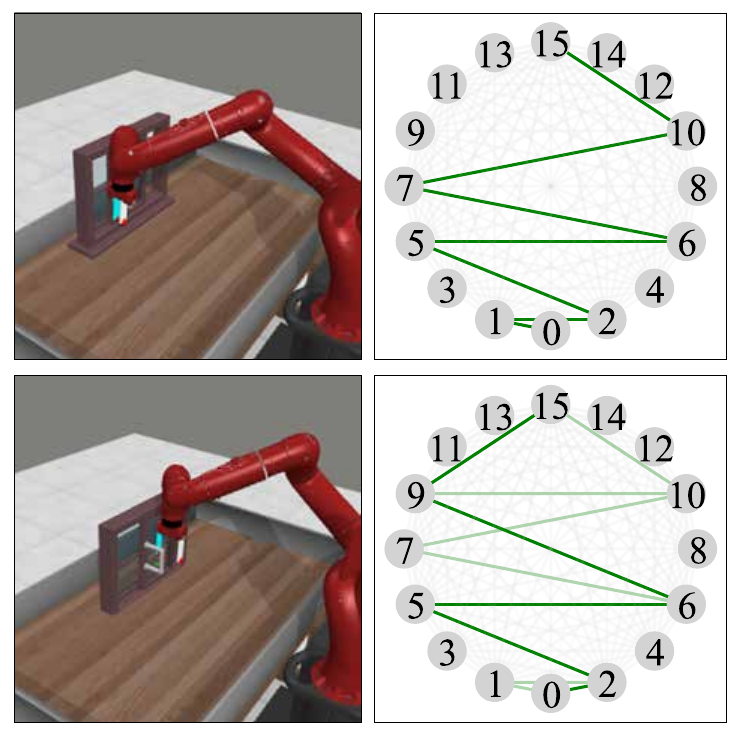}
  \caption{window-close/open}
\end{subfigure}
\caption{
(a): Visualization of 1,000 samples per task with t-SNE.
(b)-(d): Comparing routing paths of D2R with complete 16 modules.
The light gray fully connected graph displays all possible connections, while the green edges represent the routing paths, where darker color means higher routing probability.
Each column illustrates two different tasks.}
\label{fig:visualization}
\end{figure*}

\paragraph{Benchmarks.} 
We evaluate D2R on Meta-World \cite{yu2020meta}, an MTRL benchmark consisting of 50 robotics manipulation tasks with a sawyer arm in the MuJoCo environment \cite{todorov2012mujoco}.
It offers two setups: \textbf{MT10} (a suite of 10 tasks) and \textbf{MT50} (a suite of 50 tasks).
Following \cite{yang2020multi}, we extend all tasks to a random-goal setting.
Specifically, \textbf{MT10-Fixed} and \textbf{MT50-Fixed} refer to tasks with fixed goals, while \textbf{MT10-Rand} and \textbf{MT50-Rand} denote tasks with random goals.

\paragraph{Baselines.} 
We compare D2R against five baselines. 
(i) \textbf{Multi-task SAC}: Extension of SAC for MTRL with one-hot encoding for task representation.
(ii) \textbf{Multi-head SAC}: SAC with a shared network backbone apart from independent heads for each task.
(iii) \textbf{SM} \cite{yang2020multi}: Learning a routing network for soft combination of modules.
(iv) \textbf{CARE} \cite{sodhani2021multi}: Leveraging additional metadata and a mixture of encoders for task representation.
(v) \textbf{PaCo} \cite{sun2022paco}: A parameter compositional approach to recombine task-specific parameters.

\paragraph{Training Settings.}
We train all methods with 15 million environment steps (1.5 million per task) using 10 parallel environments on the MT10 setting and 50 million environment steps (1 million per task) using 50 parallel environments on the MT50 setting. 
Because effective knowledge sharing between tasks improves sampling efficiency and thus accelerates convergence, we reduce the number of sampling environment steps per task as the number of tasks increases.

\begin{figure*}[t]
\centering
\includegraphics[width=0.98\textwidth]{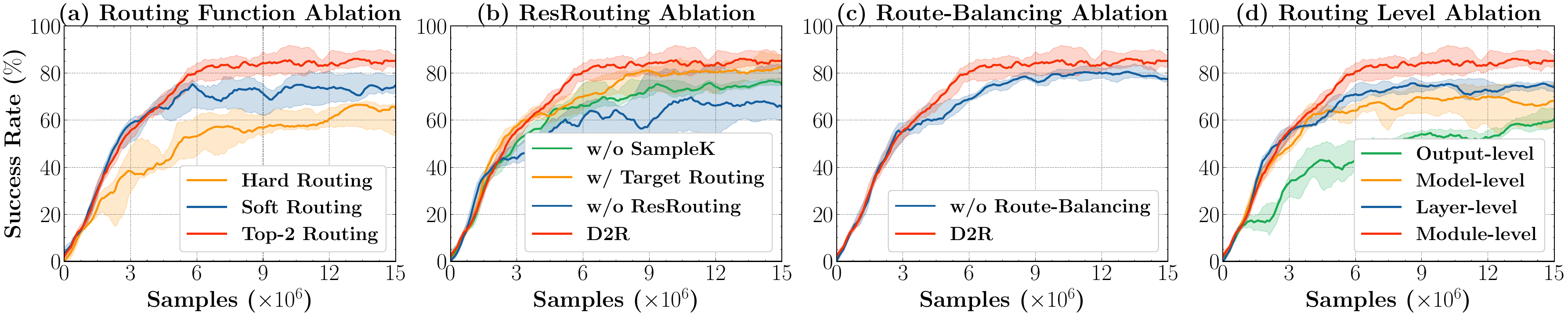}
\caption{Four distinct ablation studies on the MT10-Rand setting.}
\label{fig:result_ablation}
\end{figure*}

\begin{figure}[t]
\centering
\includegraphics[width=0.99\columnwidth]{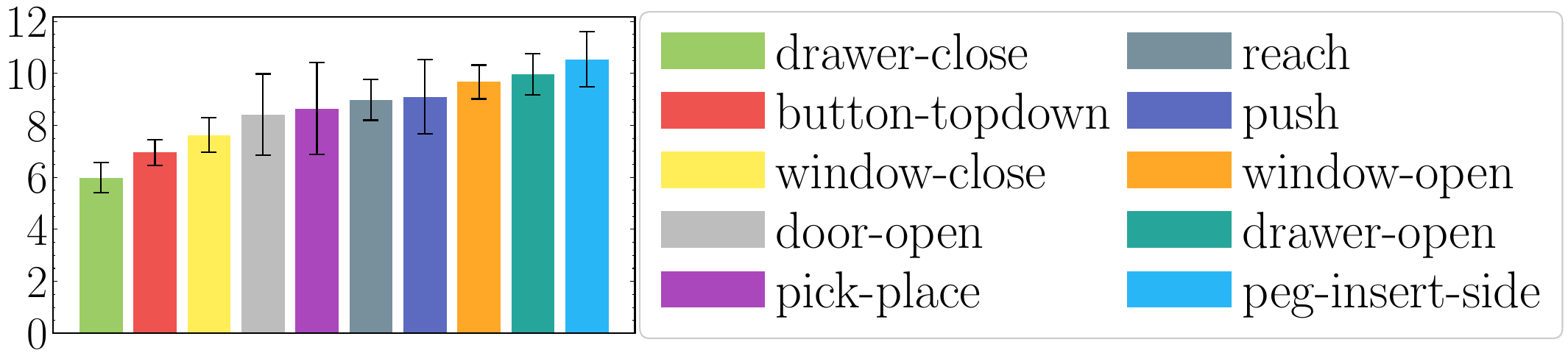}
\caption{Number of modules utilized in each task.}
\label{fig:bar-plot}
\end{figure}

\subsection{Quantitative Result} \label{sec:Quantitative Results}

The evaluation metrics of the learned universal policy are based on the success rate of task execution, which is well-defined in the Meta-World benchmark.
We evaluate each method every 10K steps per task with 32 episodes and report the 1st, median, and 3rd quartile success rates across 5 different seeds, as illustrated in Fig.\ref{fig:result}.
D2R is closely related to SM, a typical layer-level routing method. 
SM consists of 4 layers, each containing 4 modules.
Thus, we use the same setup of 16 modules to ensure a fair comparison.
In both MT10 settings, D2R exhibits impressive performance and stands out as the only method achieving nonzero successful rates across all ten tasks.
We also observe that in MT50 settings, D2R demonstrates improved success rates and faster convergence compared to the baseline methods.
For the final performance evaluation, we assess the final policy through 100 episodes per task and present the results alongside the parameter size of each method in Table.8 (Appendix C.1).

\subsection{Routing Analysis} \label{sec:Routing Analysis}
To gain a deeper understanding of D2R's behavior, we analyze its routing on MT10-Rand from three aspects.

\paragraph{t-SNE Visualization.}
We evaluate the final policy on each task five times, collecting 1,000 samples of routing paths per task.
We reduce the dimension and visualize the routing paths with t-SNE \cite{van2008visualizing} in Fig.\ref{fig:visualization}(a).
The result demonstrates clear boundaries between tasks, showcasing the agent's ability to distinguish different tasks.
Additionally, we observe intersections among certain tasks as they share common skill sets.
For instance, [\textit{push}, \textit{pick-place}, \textit{peg-insert-side}] and [\textit{window-open}, \textit{window-close}] exhibit overlaps.
Furthermore, certain tasks, such as \textit{reach}, are divided into multiple sets because they represent fundamental skills that most modules need to handle.

\paragraph{Routing Path Visualization.} 
We present a comparison of complete routing paths for different tasks in Fig.\ref{fig:visualization}(b), (c), and (d).
Each image pair illustrates the state at a specific timestep on the left side and the corresponding routing path at that moment on the right.
The comparison between task \textit{pick-place} and \textit{peg-insert-side} (Fig.\ref{fig:visualization}(b)) shows similar routing paths, as they both involve the skill of moving an object from one location to another, with the exception that \textit{pick-place} utilizes an additional module $M^4$.
In another comparison, task \textit{door-open} and \textit{drawer-open} (Fig.\ref{fig:visualization}(c)) share nearly identical routing paths during the open operation.
Likewise, task \textit{window-close} and \textit{window-open} (Fig.\ref{fig:visualization}(d)) exhibit similar routing paths. Interestingly, we find that the routing path of \textit{window-close} eventually converges to single routing connections, which does not require inter-module information fusion despite using the Top-2 routing function.

\paragraph{Routing in Relation to Task Difficulty.}
We count the number of modules utilized in 1,000 samples per task and report the mean count for each task with standard derivations in Fig.\ref{fig:bar-plot}.
It is evident that difficult tasks (\eg, \textit{peg-insert-side} uses $10.6$ modules on average) indeed use more modules than easy ones (\eg., \textit{drawer-close} uses $6.0$ on average).

\subsection{Ablation Study}

\paragraph{Effect of Top-K Routing Function.}
\label{sec:Effect of Top-K Routing Function}
Besides the Top-K Routing function we employ, D2R is also compatible with other routing functions.
We show the performance of D2R under various routing functions in Fig.\ref{fig:result_ablation}(a).
Hard Routing, also known as Top-1 Routing, selects a single routing source for each module.
To address the non-differentiable problem of single path routing, we retain the routing probability of the Softmax function for all routing sources without rescaling the selected one \cite{fedus2022switch}.
In contrast, Soft Routing establishes all potential routing connections to form a maximum DAG routing path.
Hard Routing restricts modules to single connections without module knowledge fusion information, while Soft Routing wastes computing resources on optimizing unnecessary modules.
Consequently, both routing functions perform worse than our Top-2 approach.
We find that Soft Routing also converges to a sparse path for each input.
The detailed discussion on this finding in Appendix C.3 demonstrates that Top-2 Routing is beneficial for enhancing learning efficiency by pruning 
insignificant dense routing during training.

\paragraph{Effect of SampleK Routing and ResRouting Method.}
In Fig.\ref{fig:result_ablation}(b), we investigate the impact of different routing strategies on D2R during off-policy training.
When we exclude ResRouting (D2R \without ResRouting), there is a significant decrease in the success rate and an increase in variance.
Hence, ResRouting plays an important role in mitigating negative knowledge transfer when training D2R with the behavior policy routing paths.
Since the target policy can learn from data generated by a completely unrelated behavior policy (\eg, from a human expert)~\cite{sutton2018reinforcement}, it can adopt its own routing paths instead of emulating the behavior policy.
Thus, we also test D2R \with Target Routing, which uses newly sampled routing paths. Its performance and stability are improved compared to D2R \without ResRouting.
However, it falls short of D2R because the improved routing paths explored by the behavior policy are not well-exploited in training.
Another variant, D2R \without SampleK, uses TopK (Eq.\ref{eq:topk}) instead of SampleK (Eq.\ref{eq:samplek}) during the training phase, but it proves to be detrimental to performance because it lacks exploration of various routing paths.

\paragraph{Effect of Automatic Route-Balancing Mechanism.}
We present the performance of D2R without the automatic route-balancing mechanism in Fig.\ref{fig:result_ablation}(c).
D2R \without Route-Balancing exhibits comparable learning efficiency with D2R in the initial training phase. However, as the agent becomes proficient in certain tasks, D2R with the route-balancing mechanism encourages enhanced exploration of routing paths for the tasks that are not yet mastered, thereby leading to improved overall performance.

\paragraph{Effect of Module-Level Routing.}
We reimplement both model-level and layer-level routing types by integrating all key components of D2R, \ie, the Top-K routing function, the SampleK routing exploration, the ResRouting method, and the automatic route-balancing mechanism.
The results, depicted in Fig.\ref{fig:result_ablation}(d), demonstrate that as the routing flexibility and the expressive capability improve, the corresponding method also shows improved performance.
D2R's module-level routing offers the highest flexibility and the strongest expressive capability, leading to the best performance.

\section{Conclusion and Discussion}

In this work, we start by introducing module-level routing, which enables more flexible knowledge sharing.
We then propose D2R, designed to adapt varying quantities of knowledge for tasks with differing complexity.
To address the challenge of disparate routing paths during off-policy training, we present the ResRouting method as a solution.
Furthermore, we design an automatic route-balancing mechanism to balance routing exploration and exploitation for unmastered and mastered tasks.
As a result, D2R significantly improves knowledge sharing and outperforms other baseline methods.

One limitation of our proposed approach is that the number of base modules is predetermined, potentially leading to situations where specific modules remain unused by any tasks.
Consequently, exploring methods for dynamically generating modules as required by the task presents an intriguing avenue for future research.
Furthermore, since modularization provides a natural incremental modeling structure, D2R's ability for other challenging tasks like few-shot transfer learning will also need investigations.

\section{Acknowledgements}

This work is supported in part by the National Key Research and Development Program of China under Grant 2022ZD0116401;
the Natural Science Foundation of China under Grant 62076238, Grant 62222606, and Grant 61902402;
the Jiangsu Key Research and Development Plan (No. BE2023016);
and the China Computer Federation (CCF)-Tencent Open Fund.

\bibliography{aaai24}

\begin{thebibliography}{36}
\providecommand{\natexlab}[1]{#1}

\bibitem[{Andreas, Klein, and Levine(2017)}]{andreas2017modular}
Andreas, J.; Klein, D.; and Levine, S. 2017.
\newblock Modular Multitask Reinforcement Learning with Policy Sketches.
\newblock In \emph{Proceedings of International Conference on Machine Learning}, 166--175.

\bibitem[{Bellman(1966)}]{bellman1966dynamic}
Bellman, R. 1966.
\newblock Dynamic Programming.
\newblock \emph{Science}, 153(3731): 34--37.

\bibitem[{Bolukbasi et~al.(2017)Bolukbasi, Wang, Dekel, and Saligrama}]{bolukbasi2017adaptive}
Bolukbasi, T.; Wang, J.; Dekel, O.; and Saligrama, V. 2017.
\newblock Adaptive Neural Networks for Efficient Inference.
\newblock In \emph{Proceedings of International Conference on Machine Learning}, 527--536.

\bibitem[{Calandriello, Lazaric, and Restelli(2014)}]{calandriello2014sparse}
Calandriello, D.; Lazaric, A.; and Restelli, M. 2014.
\newblock Sparse Multi-Task Reinforcement Learning.
\newblock In \emph{Advances in Neural Information Processing Systems}, 819--827.

\bibitem[{Caruana(1997)}]{caruana1997multitask}
Caruana, R. 1997.
\newblock Multitask Learning.
\newblock \emph{Machine Learning}, 28: 41--75.

\bibitem[{Cho, Jung, and Sung(2022)}]{cho2022multi}
Cho, M.; Jung, W.; and Sung, Y. 2022.
\newblock Multi-Task Reinforcement Learning with Task Representation Method.
\newblock In \emph{ICLR Workshop on Generalizable Policy Learning in Physical World}, 1--11.

\bibitem[{Fedus, Zoph, and Shazeer(2022)}]{fedus2022switch}
Fedus, W.; Zoph, B.; and Shazeer, N. 2022.
\newblock Switch Transformers: Scaling to Trillion Parameter Models with Simple and Efficient Sparsity.
\newblock \emph{Journal of Machine Learning Research}, 23(1): 5232--5270.

\bibitem[{Fernando et~al.(2017)Fernando, Banarse, Blundell, Zwols, Ha, Rusu, Pritzel, and Wierstra}]{fernando2017pathnet}
Fernando, C.; Banarse, D.; Blundell, C.; Zwols, Y.; Ha, D.; Rusu, A.~A.; Pritzel, A.; and Wierstra, D. 2017.
\newblock PathNet: Evolution Channels Gradient Descent in Super Neural Networks.
\newblock arXiv:1701.08734.

\bibitem[{Haarnoja et~al.(2018)Haarnoja, Zhou, Abbeel, and Levine}]{haarnoja2018soft}
Haarnoja, T.; Zhou, A.; Abbeel, P.; and Levine, S. 2018.
\newblock Soft Actor-Critic: Off-policy Maximum Entropy Deep Reinforcement Learning with a Stochastic Actor.
\newblock In \emph{Proceedings of International Conference on Machine Learning}, 1856--1865.

\bibitem[{He et~al.(2016)He, Zhang, Ren, and Sun}]{he2016deep}
He, K.; Zhang, X.; Ren, S.; and Sun, J. 2016.
\newblock Deep Residual Learning for Image Recognition.
\newblock In \emph{Proceedings of the IEEE Conference on Computer Vision and Pattern Recognition}, 770--778.

\bibitem[{Hessel et~al.(2019)Hessel, Soyer, Espeholt, Czarnecki, Schmitt, and Van~Hasselt}]{hessel2019multi}
Hessel, M.; Soyer, H.; Espeholt, L.; Czarnecki, W.; Schmitt, S.; and Van~Hasselt, H. 2019.
\newblock Multi-Task Deep Reinforcement Learning with Popart.
\newblock In \emph{Proceedings of the AAAI Conference on Artificial Intelligence}, 3796--3803.

\bibitem[{Jacobs et~al.(1991)Jacobs, Jordan, Nowlan, and Hinton}]{jacobs1991adaptive}
Jacobs, R.~A.; Jordan, M.~I.; Nowlan, S.~J.; and Hinton, G.~E. 1991.
\newblock Adaptive Mixtures of Local Experts.
\newblock \emph{Neural Computation}, 3(1): 79--87.

\bibitem[{Lepikhin et~al.(2021)Lepikhin, Lee, Xu, Chen, Firat, Huang, Krikun, Shazeer, and Chen}]{lepikhin2020gshard}
Lepikhin, D.; Lee, H.; Xu, Y.; Chen, D.; Firat, O.; Huang, Y.; Krikun, M.; Shazeer, N.; and Chen, Z. 2021.
\newblock GShard: Scaling Giant Models with Conditional Computation and Automatic Sharding.
\newblock In \emph{Proceedings of International Conference on Learning Representations}, 1--14.

\bibitem[{Levine et~al.(2016)Levine, Finn, Darrell, and Abbeel}]{levine2016end}
Levine, S.; Finn, C.; Darrell, T.; and Abbeel, P. 2016.
\newblock End-to-end Training of Deep Visuomotor Policies.
\newblock \emph{The Journal of Machine Learning Research}, 17(1): 1334--1373.

\bibitem[{Lillicrap et~al.(2016)Lillicrap, Hunt, Pritzel, Heess, Erez, Tassa, Silver, and Wierstra}]{lillicrap2015continuous}
Lillicrap, T.~P.; Hunt, J.~J.; Pritzel, A.; Heess, N.; Erez, T.; Tassa, Y.; Silver, D.; and Wierstra, D. 2016.
\newblock Continuous Control with Deep Reinforcement Learning.
\newblock In \emph{Proceedings of International Conference on Learning Representations}, 1--10.

\bibitem[{Liu et~al.(2021)Liu, Liu, Jin, Stone, and Liu}]{liu2021conflict}
Liu, B.; Liu, X.; Jin, X.; Stone, P.; and Liu, Q. 2021.
\newblock Conflict-Averse Gradient Descent for Multi-Task Learning.
\newblock In \emph{Advances in Neural Information Processing Systems}, 18878--18890.

\bibitem[{Ma et~al.(2018)Ma, Zhao, Yi, Chen, Hong, and Chi}]{ma2018modeling}
Ma, J.; Zhao, Z.; Yi, X.; Chen, J.; Hong, L.; and Chi, E.~H. 2018.
\newblock Modeling Task Relationships in Multi-Task Learning with Multi-Gate Mixture-of-Experts.
\newblock In \emph{Proceedings of ACM SIGKDD International Conference on Knowledge Discovery \& Data Mining}, 1930–1939.

\bibitem[{Mnih et~al.(2015)Mnih, Kavukcuoglu, Silver, Rusu, Veness, Bellemare, Graves, Riedmiller, Fidjeland, Ostrovski, Petersen, Beattie, Sadik, Antonoglou, King, Kumaran, Wierstra, Legg, and Hassabis}]{mnih2015human}
Mnih, V.; Kavukcuoglu, K.; Silver, D.; Rusu, A.~A.; Veness, J.; Bellemare, M.~G.; Graves, A.; Riedmiller, M.~A.; Fidjeland, A.; Ostrovski, G.; Petersen, S.; Beattie, C.; Sadik, A.; Antonoglou, I.; King, H.; Kumaran, D.; Wierstra, D.; Legg, S.; and Hassabis, D. 2015.
\newblock Human-Level Control through Deep Reinforcement Learning.
\newblock \emph{Nature}, 518(7540): 529--533.

\bibitem[{Parisotto, Ba, and Salakhutdinov(2016)}]{parisotto2015actor}
Parisotto, E.; Ba, L.~J.; and Salakhutdinov, R. 2016.
\newblock Actor-Mimic: Deep Multitask and Transfer Reinforcement Learning.
\newblock In \emph{Proceedings of International Conference on Learning Representations}, 1--9.

\bibitem[{Puterman(2014)}]{puterman2014markov}
Puterman, M.~L. 2014.
\newblock \emph{Markov Decision Processes: Discrete Stochastic Dynamic Programming}.
\newblock John Wiley \& Sons.

\bibitem[{Shazeer et~al.(2017)Shazeer, Mirhoseini, Maziarz, Davis, Le, Hinton, and Dean}]{shazeer2017outrageously}
Shazeer, N.; Mirhoseini, A.; Maziarz, K.; Davis, A.; Le, Q.~V.; Hinton, G.~E.; and Dean, J. 2017.
\newblock Outrageously Large Neural Networks: The Sparsely-Gated Mixture-of-Experts Layer.
\newblock In \emph{Proceedings of International Conference on Learning Representations}, 1--12.

\bibitem[{Singh(1992)}]{singh1992transfer}
Singh, S.~P. 1992.
\newblock Transfer of Learning by Composing Solutions of Elemental Sequential Tasks.
\newblock \emph{Machine Learning}, 8: 323--339.

\bibitem[{Sodhani, Zhang, and Pineau(2021)}]{sodhani2021multi}
Sodhani, S.; Zhang, A.; and Pineau, J. 2021.
\newblock Multi-Task Reinforcement Learning with Context-based Representations.
\newblock In \emph{Proceedings of International Conference on Machine Learning}, 9767--9779.

\bibitem[{Sun et~al.(2022)Sun, Zhang, Xu, and Tomizuka}]{sun2022paco}
Sun, L.; Zhang, H.; Xu, W.; and Tomizuka, M. 2022.
\newblock PaCo: Parameter-Compositional Multi-Task Reinforcement Learning.
\newblock In \emph{Advances in Neural Information Processing Systems}, 21495--21507.

\bibitem[{Sutton and Barto(2018)}]{sutton2018reinforcement}
Sutton, R.~S.; and Barto, A.~G. 2018.
\newblock \emph{Reinforcement Learning: An Introduction}.
\newblock MIT press.

\bibitem[{Teerapittayanon, McDanel, and Kung(2016)}]{teerapittayanon2016branchynet}
Teerapittayanon, S.; McDanel, B.; and Kung, H. 2016.
\newblock BranchyNet: Fast Inference via Early Exiting from Deep Neural Networks.
\newblock In \emph{Proceedings of International Conference on Pattern Recognition}, 2464--2469.

\bibitem[{Teh et~al.(2017)Teh, Bapst, Czarnecki, Quan, Kirkpatrick, Hadsell, Heess, and Pascanu}]{teh2017distral}
Teh, Y.; Bapst, V.; Czarnecki, W.~M.; Quan, J.; Kirkpatrick, J.; Hadsell, R.; Heess, N.; and Pascanu, R. 2017.
\newblock Distral: Robust Multitask Reinforcement Learning.
\newblock In \emph{Advances in Neural Information Processing Systems}, 4496--4506.

\bibitem[{Todorov, Erez, and Tassa(2012)}]{todorov2012mujoco}
Todorov, E.; Erez, T.; and Tassa, Y. 2012.
\newblock Mujoco: A Physics Engine for Model-based Control.
\newblock In \emph{Proceedings of IEEE/RSJ International Conference on Intelligent Robots and Systems}, 5026--5033.

\bibitem[{Van~der Maaten and Hinton(2008)}]{van2008visualizing}
Van~der Maaten, L.; and Hinton, G. 2008.
\newblock Visualizing Data using t-SNE.
\newblock \emph{Journal of Machine Learning Research}, 9(86): 2579--2605.

\bibitem[{Veit and Belongie(2018)}]{veit2018convolutional}
Veit, A.; and Belongie, S. 2018.
\newblock Convolutional Networks with Adaptive Inference Graphs.
\newblock In \emph{Proceedings of the European Conference on Computer Vision}, 3--18.

\bibitem[{Wang et~al.(2018)Wang, Yu, Dou, Darrell, and Gonzalez}]{wang2018skipnet}
Wang, X.; Yu, F.; Dou, Z.-Y.; Darrell, T.; and Gonzalez, J.~E. 2018.
\newblock SkipNet: Learning Dynamic Routing in Convolutional Networks.
\newblock In \emph{Proceedings of the European Conference on Computer Vision}, 409--424.

\bibitem[{Xu et~al.(2020)Xu, Wu, Che, Tang, and Ye}]{xu2020knowledge}
Xu, Z.; Wu, K.; Che, Z.; Tang, J.; and Ye, J. 2020.
\newblock Knowledge Transfer in Multi-Task Deep Reinforcement Learning for Continuous Control.
\newblock In \emph{Advances in Neural Information Processing Systems}, 15146--15155.

\bibitem[{Yang et~al.(2020)Yang, Xu, Wu, and Wang}]{yang2020multi}
Yang, R.; Xu, H.; Wu, Y.; and Wang, X. 2020.
\newblock Multi-Task Reinforcement Learning with Soft Modularization.
\newblock In \emph{Advances in Neural Information Processing Systems}, 4767--4777.

\bibitem[{Ye et~al.(2020)Ye, Liu, Sun, Shi, Zhao, Wu, Yu, Yang, Wu, Guo, Chen, Yin, Zhang, Shi, Wang, Fu, Yang, and Huang}]{ye2020mastering}
Ye, D.; Liu, Z.; Sun, M.; Shi, B.; Zhao, P.; Wu, H.; Yu, H.; Yang, S.; Wu, X.; Guo, Q.; Chen, Q.; Yin, Y.; Zhang, H.; Shi, T.; Wang, L.; Fu, Q.; Yang, W.; and Huang, L. 2020.
\newblock Mastering Complex Control in MOBA Games with Deep Reinforcement Learning.
\newblock In \emph{Proceedings of the AAAI Conference on Artificial Intelligence}, 6672--6679.

\bibitem[{Yu et~al.(2020{\natexlab{a}})Yu, Kumar, Gupta, Levine, Hausman, and Finn}]{yu2020gradient}
Yu, T.; Kumar, S.; Gupta, A.; Levine, S.; Hausman, K.; and Finn, C. 2020{\natexlab{a}}.
\newblock Gradient Surgery for Multi-Task Learning.
\newblock In \emph{Advances in Neural Information Processing Systems}, 5824--5836.

\bibitem[{Yu et~al.(2020{\natexlab{b}})Yu, Quillen, He, Julian, Hausman, Finn, and Levine}]{yu2020meta}
Yu, T.; Quillen, D.; He, Z.; Julian, R.; Hausman, K.; Finn, C.; and Levine, S. 2020{\natexlab{b}}.
\newblock Meta-World: A Benchmark and Evaluation for Multi-Task and Meta Reinforcement Learning.
\newblock In \emph{Proceedings of the Conference on Robot Learning}, 1094--1100.

\end{thebibliography}

\newpage
$$$$

\newpage

\appendix

\begin{table}[t]
\centering
\begin{tabularx}{\linewidth}{Xc}
\toprule
\makecell[c]{Hyperparameter} & Hyperparameter Values \\
\midrule
network architecture        & feedforward network \\
batch size                  & 128 $\times$ number of tasks \\
non-linearity               & ReLU \\
policy initialization       & standard Gaussian \\
\# of samples / \# of train steps per iteration   & 1 env step / 1 training step \\
policy learning rate        & 3e-4 \\
Q function learning rate    & 3e-4 \\
optimizer                   & Adam  \\
discount                    & .99 \\
episode length              & 200 \\
exploration steps           & 2000 $\times$ number of tasks \\
reward scale                & 0.1 \\
replay buffer size          & 1e6 (MT10) / 1e7 (MT50) \\
\bottomrule
\end{tabularx}
\caption{Common hyperparameters across all methods.}
\label{tb:common-hyper}
\end{table}

\begin{table}[t]
\centering
\begin{tabularx}{\linewidth}{Xc}
\toprule
\makecell[c]{Hyperparameter} & Hyperparameter Values \\
\midrule
network size                & [400, 400, 400, 400, 400] \\
\bottomrule
\end{tabularx}
\caption{Additional hyperparameters for Multi-task SAC.}
\label{tb:mt-hyper}
\end{table}

\begin{table}[h!]
\centering
\begin{tabularx}{\linewidth}{Xc}
\toprule
\makecell[c]{Hyperparameter} & Hyperparameter Values \\
\midrule
network architecture        & multi-head (1 head/task) \\
network size                & [400, 400, 400, 400, 400] \\
\bottomrule
\end{tabularx}
\caption{Additional hyperparameters for Multi-head SAC.}
\end{table}

\begin{table}[h!]
\centering
\begin{tabularx}{\linewidth}{Xc}
\toprule
\makecell[c]{Hyperparameter} & Hyperparameter Values \\
\midrule
state representation size   & [400, 400, 400] \\
task representation size    & [128, 128] \\
number of encoders          & 6 \\
encoder gating size         & [128, 128] \\
encoder network size        & [128, 128] \\
\bottomrule
\end{tabularx}
\caption{Additional hyperparameters for CARE.}
\end{table}

\begin{table}[h!]
\centering
\begin{tabularx}{\linewidth}{Xc}
\toprule
\makecell[c]{Hyperparameter} & Hyperparameter Values \\
\midrule
state representation size   & [400, 400] \\
task representation size    & [400] \\
number of layers            & 4 \\
number of modules/layer     & 4 \\
module size                 & 128 \\
routing size                & [256, 256, 256, 256, 256] \\
\bottomrule
\end{tabularx}
\caption{Additional hyperparameters for SM.}
\end{table}

\begin{table}[h!]
\centering
\begin{tabularx}{\linewidth}{Xc}
\toprule
\makecell[c]{Hyperparameter} & Hyperparameter Values \\
\midrule
param-set size              & [400, 400, 400] \\
number pf param-sets        & 5 \\   
extreme loss threshold      & 3e3 \\
\bottomrule
\end{tabularx}
\caption{Additional hyperparameters for PaCo.}
\label{tb:paco-hyper}
\end{table}

\begin{table}[h!]
\centering
\begin{tabularx}{\linewidth}{Xc}
\toprule
\makecell[c]{Hyperparameter} & Hyperparameter Values \\
\midrule
state representation size   & [400, 400] \\
task representation size    & [400] \\
number of modules           & 16 \\
module size                 & 128 \\
routing size                & [256, 256] \\
\bottomrule
\end{tabularx}
\caption{Additional hyperparameters for D2R.}
\label{tb:d2r-hyper}
\end{table}

\section{Additional Implementation Details} \label{sec:Additional Implementation Details}
\subsection{Libraries}

For baselines, we used the Soft-Module\footnote{https://github.com/RchalYang/Soft-Module} codebase with an extra modification using AsyncVectorEnv class implemented by Gym\footnote{https://github.com/openai/gym}, which enables us to run our experiments across multiple parallel environments.
Unlike \cite{yang2020multi, sodhani2021multi}, we run our experiments on the Meta-World-V2 benchmark\footnote{https://github.com/Farama-Foundation/Metaworld}.

\begin{algorithm*}[tb]
\caption{Dynamic Depth Routing (D2R)}
\label{alg:d2r}
\textbf{Input}: number of modules $N$, routing Top-K value $K$\\
\textbf{Initialization}: routing network $G_\phi$, base module network $M_\phi$, replay buffer $\mathcal{D} \leftarrow \emptyset$
\begin{algorithmic}[1] 

\For{each epoch $j=1,2,\dots$}
    \State $//$ Sampling: 
    \For{each task $\mathcal{T}$}
        \State Compute temperature coefficient in Eq.\ref{eq:temp} as $\tau_\mathcal{T} = \operatorname{Softmax}(-\log(\alpha_\mathcal{T}))$ \Comment{Routing Exploration Balance}
        \For{each timestep $t=1,2,\dots,T$}
            \For{each module $i = 1,2,\dots,N$} \Comment{Dynamic Depth Routing}
                \State Compute routing logits in Eq.\ref{eq:routing} as $z^i = G_\phi^i (F_\phi (s_t) \cdot H_\phi(\mathcal{T}))$
                \State Compute routing paths in Eq.\ref{eq:temp-samplek} as $d^i = \operatorname{SampleK}\left( z^i/\tau_\mathcal{T}, K \right)$
                \State Compute routing probabilities in Eq.\ref{eq:maskedsoftmax} as $p^i = \operatorname{MaskSoftmax}\left(z^i, d^i\right)$
                \If{$i < N$}
                    \State Compute module output $m^i = \sum_{j=1}^{i-1} p^i_j \cdot {m}^j + M_\phi^i\left(\sum_{j=1}^{i-1} p^i_j \cdot {m}^j\right)$
                \Else
                    \State Compute action distribution $\pi_\mathcal{T} = M_\phi^{i}\left(\sum_{j=1}^i p^i_j \cdot {m}^j\right)$
                \EndIf
            \EndFor
            \State Sample action $a_t \sim \pi_\mathcal{T}(\cdot|s_t, z_\mathcal{T})$ and execute $a_t$
            \State Get current state $s_{t+1}$ and reward $r_t \leftarrow R(s_t, a_t, \mathcal{T})$
            \State Add $\langle s_t, a_t, r_t, s_{t+1}, d \rangle$ to $\mathcal{D}$
        \EndFor
    \EndFor
    \State $//$ Training: 
    \For{each training step $t=1,2,\dots,T$}
        \State Sample a mini-batch $\mathcal{S} = \langle s_t, a_t, r_t, s_{t+1}, d_{\text{old}} \rangle$ from $\mathcal{D}$
        \For{each module $i = 1,2,\dots,N$}
            \State Compute routing logits $z^i$ as in line 7
            \State Compute routing probabilities using $d_{\text{old}}$ in Eq.\ref{eq:keepk-maskedsoftmax} as $p^i = \operatorname{MaskSoftmax}\left(z^i, d_{\text{old}}^i\right)$ \Comment{ResRouting}
            \If{$i < N$}
                \State Compute module output in Eq.\ref{eq:resrouting3} as $m^i = \sum_{j=1}^{i-1} p^i_j \cdot \chi\left(m^j, z^i_j, \operatorname{rsg}\right) + M_\phi^{i}  \left(\sum_{j=1}^{i-1} p^i_j \cdot \chi\left(m^j, z^i_j, \operatorname{rsg}\right) \right)$
            \Else
                \State Compute action distribution and Q-value as in line 13
            \EndIf
        \EndFor
        \State Update $\phi$ with SAC loss
    \EndFor
\EndFor
\end{algorithmic}
\end{algorithm*}

\subsection{Hyperparameter Details}
In this section, we provide the detailed hyperparameter values for each method in our experiment.
General hyperparameters, shared by D2R and all baselines, are illustrated in Table.\ref{tb:common-hyper}.
Furthermore, Table.\ref{tb:mt-hyper} to Table.\ref{tb:paco-hyper} show additional hyperparameters specific to each baseline.
Eventually, Table.\ref{tb:d2r-hyper} provides the additional hyperparameters of D2R.

It is worth mentioning that we set the episode length to 200 timesteps, different from the previous setting of 150.
We evaluate the rule-based policies provided by the Meta-World benchmark~\cite{yu2020meta} with 100 sample episodes.
Under the initial episode length setting of 150 timesteps, 42 tasks get a success rate exceeding $90\%$; however, among the remaining 8 tasks, the task \textit{disassemble} exhibits a mere $50\%$ success rate.
In contrast, when we adjust the episode length setting to 200 timesteps, the success rates for all 50 tasks are beyond $90\%$, with the task \textit{disassemble} achieving an improved success rate of $91\%$.

\begin{table*}[t]
\centering
\begin{tabular}{lccccc}
\toprule
\makecell[c]{Methods} & Params (MT10/MT50) & MT10-Fixed & MT10-Rand & MT50-Fixed & MT50-Rand \\
\midrule
Multi-task SAC  & 1.33M / 1.36M & 0.66 $\pm$ 0.08 & 0.51 $\pm$ 0.08 & 0.44 $\pm$ 0.03 & 0.44 $\pm$ 0.10 \\
Multi-head SAC  & 1.36M / 1.54M & 0.74 $\pm$ 0.08 & 0.60 $\pm$ 0.04 & 0.48 $\pm$ 0.06 & 0.43 $\pm$ 0.02 \\
CARE            & 1.23M / 1.26M & 0.78 $\pm$ 0.10 & 0.67 $\pm$ 0.03 & 0.49 $\pm$ 0.02 & 0.46 $\pm$ 0.03 \\
SM              & 1.59M / 1.62M & 0.84 $\pm$ 0.05 & 0.72 $\pm$ 0.03 & 0.55 $\pm$ 0.03 & 0.53 $\pm$ 0.07 \\
PaCo            & 3.39M / 3.39M & 0.80 $\pm$ 0.17 & 0.76 $\pm$ 0.04 & 0.60 $\pm$ 0.03 & 0.54 $\pm$ 0.03 \\
\midrule
D2R(Ours)       & 1.37M / 1.40M & \textbf{0.92 $\pm$ 0.07} & \textbf{0.86 $\pm$ 0.05} & \textbf{0.61 $\pm$ 0.03} & \textbf{0.63 $\pm$ 0.02} \\
\bottomrule
\end{tabular}
\caption{Comparisons on average success rates and parameter size of final policy on four benchmark settings.}
\label{tb:main-result}
\end{table*}

\begin{figure*}[t]
\centering
\includegraphics[width=0.98\textwidth]{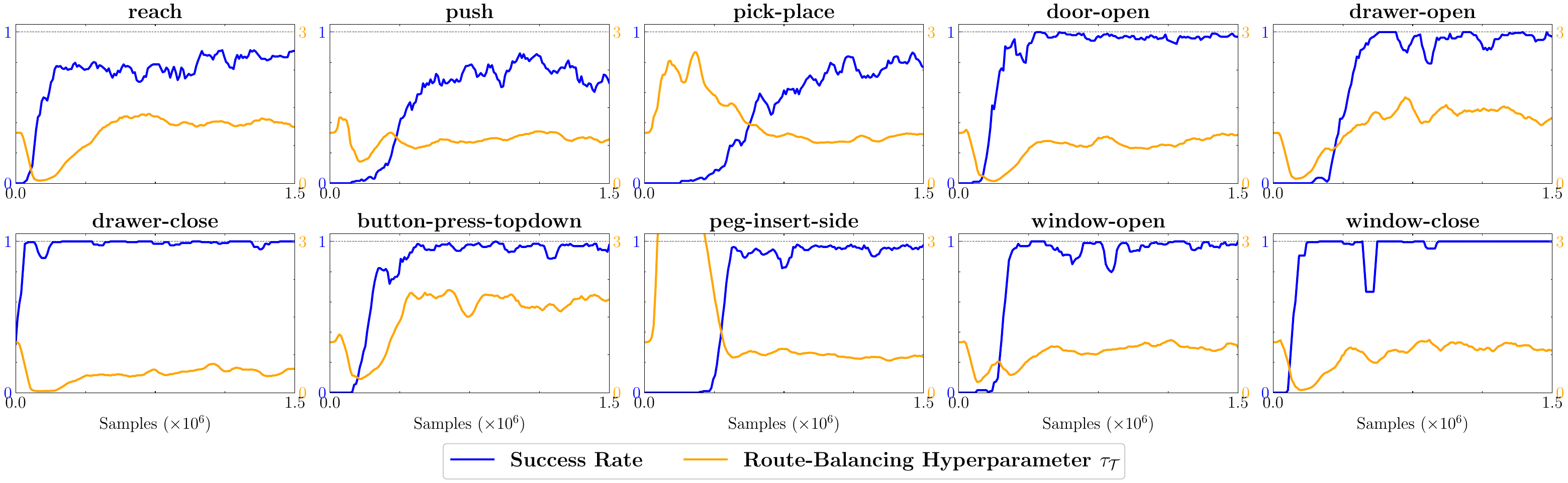}
\caption{Relationship between route-balancing hyperparameter $\tau_\mathcal{T}$ and task success rate on the MT10-Rand setting.}
\label{fig:temp-result}
\end{figure*}

In addition, for a fair comparison, we use the following tricks on all methods:

a) Disentangled SAC temperature parameters learned with different tasks.
It is a standard setup established in MTRL.
Given that distinct tasks might exhibit diverse learning dynamics throughout the training process, utilizing distinct SAC temperature parameters aids in balancing the exploration and exploitation of each task separately.

b) Loss maskout of extreme tasks.
It is utilized by \cite{sodhani2021multi, sun2022paco}, aiming to mask out the exploding loss $J_\mathcal{T}$ of task $\mathcal{T}$ from the total loss $J$ to avoid its adverse impacts on other tasks.
Specifically, once a task loss $J_\mathcal{T}$ surpasses the preconfigured threshold $\epsilon$ (set as $3e3$, the same as \cite{sun2022paco}), the task $\mathcal{T}$ will be excluded from the training loss.

c) Multi-task loss rescaling.
Cause easier tasks usually converge faster than the harder ones, to balance the training process of different tasks, \cite{yang2020multi} propose an optimization objective weight of task $\mathcal{T}$ as:
\begin{equation}
w_\mathcal{T} = \frac{\exp(-\alpha_\mathcal{T})}{\sum_{i=1}^{\mathcal{T}_N} \exp(-\alpha_i)},
\end{equation}
where $\alpha_\mathcal{T}$ is the SAC temperature parameter of task $\mathcal{T}$, and $\mathcal{T}_N$ is the total number of tasks.
So the total loss is modified as $J = \mathbb{E}_{\mathcal{T} \sim p(\mathcal{T})}[w_\mathcal{T} \cdot J_\mathcal{T}]$.

\section{Pseudo Code} \label{sec:Pseudo Code}

The pseudo-code of D2R is described in Algorithm \ref{alg:d2r}.

\section{Additional Result}

\subsection{Main Result} \label{sec:Additional Result/Main Result}
In addition to presenting the training curves comparing D2R against baselines (Section~\ref{sec:Quantitative Results}), we report the mean performance together with standard derivations of the final policies on all four benchmark settings, as summarized in Table.\ref{tb:main-result}.
Each method is trained with 5 different seeds, and each final policy is evaluated through 100 sample episodes. 
Furthermore, we compare the network parameter sizes utilized by different methods.
Among the four benchmark settings, D2R showcases the most exceptional performance.
It is worth mentioning that D2R achieves better success rates on MT50-Rand than MT50-Fixed.
It can be attributed to the fact that MT50-Rand offers diverse training examples, consequently leading to improved generalization.

\subsection{Curve for Route-Balancing Hyperparameter}

To effectively illustrate how the route-balancing hyperparameter $\tau_\mathcal{T}$ encourages exploration of routing paths for unmastered tasks while maintaining the exploitation of routing paths for mastered ones, we plot the variations of this hyperparameter value alongside the success rates of individual tasks on the MT10-Rand setting.
As shown in Fig.\ref{fig:temp-result}, the hyperparameter gradually increases when the success rate stays low, and conversely, it decreases when the success rate rises.
Notably, the route-balancing hyperparameter of the task \textit{button-press-topdown} remains high when all tasks get a high success rate since its cumulative reward per episode is relatively small compared to other tasks.
Therefore, the automatic route-balancing mechanism may work more stably by normalizing the cumulative reward across all tasks.

\begin{figure}[t]
\centering
\includegraphics[width=0.95\columnwidth]{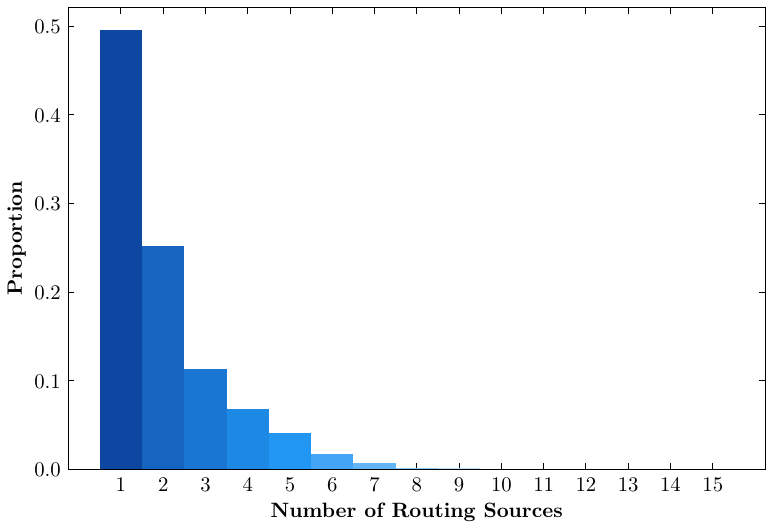}
\caption{The sparsity of D2R utilizing the soft routing function. Routing sources refer to the source modules with routing probabilities greater than 0.01.}
\label{fig:result_soft}
\end{figure}

\begin{figure}[t]
\centering
\includegraphics[width=0.95\columnwidth]{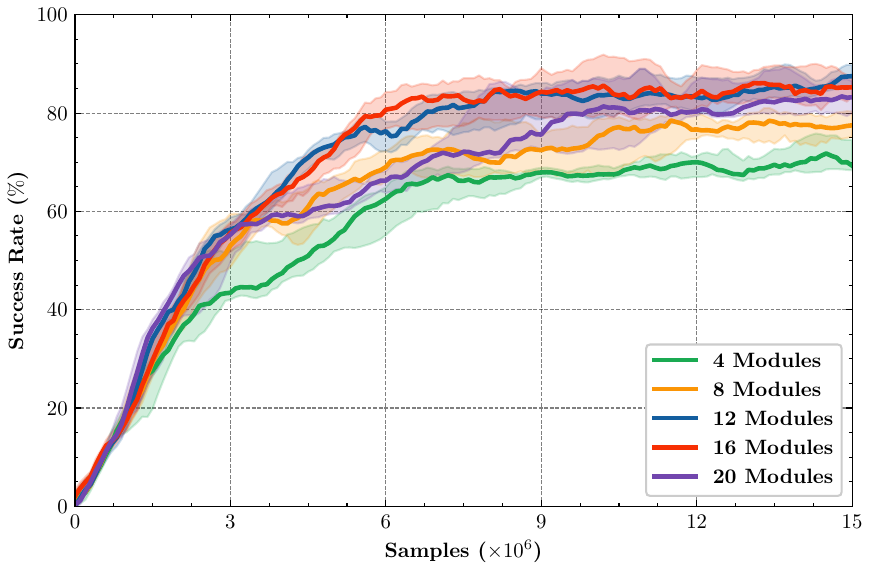}
\caption{Ablation on the number of modules.}
\label{fig:result_module_num}
\end{figure}

\subsection{Analysis of D2R with Soft Routing Function} \label{sec:Additional Result/Analysis of D2R with Soft Routing Function}

In Section \ref{sec:Effect of Top-K Routing Function}, we present a comparative analysis of D2R utilizing various routing functions, among which the top-2 routing function stands out as the most optimal choice.
We evaluate the final policy of D2R using the soft routing function with 100 sample episodes per task.
In addition, we analyze the routing probabilities of all modules in a total of $200,000$ samples ($10$ tasks $\times$ $100$ episodes $\times$ $200$ episode length).
As shown in Fig.\ref{fig:result_soft}, we represent the percentage distribution of modules with varying numbers of routing sources.
The result shows that, as training proceeds, the soft routing function gradually converges towards utilizing different sparse paths for different inputs rather than highly overlapping dense paths for all inputs.
Consequently, the top-2 routing function proves beneficial in accelerating training through pruning insignificant routing sources.

\subsection{Ablation Study on Module Quantity}

The number of modules is also a hyperparameter within this work.
SM~\cite{yang2020multi}, a representative layer-level routing method, consists of 4 layers, each containing 4 modules.
In order to ensure a fair comparison, D2R maintains the same configuration of 16 modules.
To thoroughly explore the impact of the number of modules, we illustrate D2R's performance across varying module quantities in Fig.\ref{fig:result_module_num} on the MT10-rand setting.
Our evaluation encompasses five distinct module quantities: a 4-module model, an 8-module model, a 12-module model, a 16-module model (D2R), and a 20-module model.
The result shows a significant performance improvement with increased module quantity when the number of modules is below 12.
However, once the module quantity surpasses 12, further improvements in performance become increasingly elusive, and instead, the convergence rate gradually slows down.
Therefore, on the MT10-Rand setting, an approximate total of 12 modules emerges as the optimal configuration, allowing the flexibility to construct reasonable routings for varied tasks and states.

\begin{figure}[t]
\centering
\includegraphics[width=0.95\columnwidth]{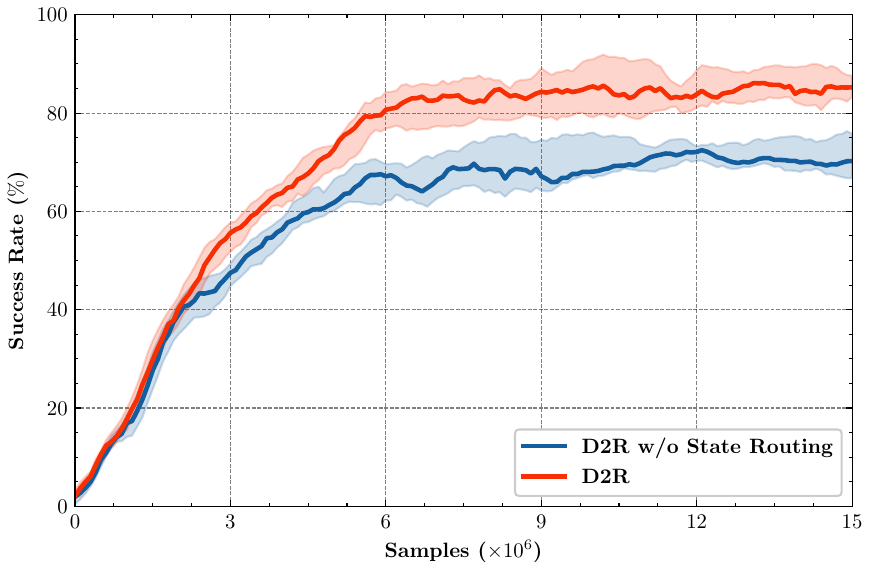}
\caption{Ablation on the routing input.}
\label{fig:result_routing_obs}
\end{figure}

\subsection{Ablation Study on Routing Input}

The original input to the routing network in D2R is the element-wise multiplication of the state representation $F(s_t)$ and the task representation $H(\mathcal{T})$.
We further present the performance of D2R \without State Routing by removing state inputs for the routing network on the MT10-Rand setting.
As shown in Fig.\ref{fig:result_routing_obs}, the task-specific routing strategy without state information falls short of D2R because it only shares knowledge at the task level, limiting the effective reuse of knowledge.
As the analysis for the routing paths of different tasks at specific timesteps in Section \ref{sec:Routing Analysis}, routing with state information allows a better understanding of knowledge sharing between diverse tasks in certain similar states, leading to improved performance.

\end{document}